\documentclass{article}

\usepackage{PRIMEarxiv}

\usepackage[backref]{hyperref} 
\hypersetup{    
hidelinks,
colorlinks=true,
linkcolor=black,
citecolor=black,
urlcolor=blue
}

\usepackage{xcolor}         
\usepackage{multirow}       
\usepackage[normalem]{ulem} 
\usepackage[utf8]{inputenc} 
\usepackage[T1]{fontenc}    
\usepackage{hyperref}       
\usepackage{url}            
\usepackage{booktabs}       
\usepackage{amsfonts}       
\usepackage{nicefrac}       
\usepackage{microtype}      
\usepackage{lipsum}
\usepackage{fancyhdr}       
\usepackage{graphicx}       
\usepackage{amsmath}        

\graphicspath{{media/}}     

\title{RoCar: A Relationship Network-based Evaluation Method for Large Language Models
\thanks{\textit{\underline{Citation}}: 
\textbf{Ming Wang et. al.. RoCar: A Relationship Network-based Evaluation Method for Large Language Models.}} 
}

\author{
  Ming Wang \\
  School of Computer Science and Engineering \\
  Northeastern University \\
  Shenyang \\
  \texttt{sci.m.wang@gmail.com}
  \And
  Wenfang Wu, Chongyun Gao, Daling Wang*, Shi Feng, Yifei Zhang \\
  School of Computer Science and Engineering \\
  Northeastern University \\
  Shenyang\\
}

\begin{document}
\maketitle

\begin{abstract}
Large language models (LLMs) have received increasing attention. However, due to the complexity of its capabilities, how to rationally evaluate the capabilities of LLMs is still a task to be solved. We propose the RoCar method, which utilizes the defined basic schemas to randomly construct a task graph and generates natural language evaluation tasks based on the task graph to evaluate the reasoning and memory abilities of LLMs respectively. Due to the very large randomness of the task construction process, it is possible to ensure that none of the LLMs to be tested has directly learned the evaluation tasks, guaranteeing the fairness of the evaluation method.
\end{abstract}


\section{Introduction}
Pre-trained Models have become the dominant approach in the field of deep learning since Transformer \cite{vaswani_attention_2017}. Buy now, the Large Language Models (LLMs) represented by ChatGPT \cite{noauthor_introducing_nodate} have received the widest attention from researchers in the field of Artificial Intelligence (AI), especially Natural Language Processing (NLP). Like LLaMA \cite{touvron2023llama}, many open-source LLMs \cite{touvron_llama_nodate, du2022glm, zeng2022glm, touvron2023llama, sun2023moss, 2023internlm} have been published. Due to the strong reasoning, generative and memory abilities acquired by LLMs during training, they are able to operate a variety of traditional tasks based on specific prompts and achieve great performance. As a result, LLMs have gained widespread interest and applications, such as in the financial \cite{Cornucopia-LLaMA-Fin-Chinese}, emotional \cite{zhang2023PICA, chen2023soulchat}, legal \cite{cui_chatlaw_2023}, medical \cite{xiong2023doctorglm, wang2023huatuo, chen2023bianque1} and educational \cite{Taoli-LLama} fields.

To evaluate the capability of LLMs and to guide the selection of more appropriate LLMs in applications, many evaluation approaches \cite{chang_survey_2023} for LLMs have been proposed by researchers. C-Eval \cite{huang2023ceval} constructed a reasoning test set of 13,948 questions in 52 subjects ranging from junior school to postgraduate university and vocational exams to evaluate LLM's problem-solving skills. Gaokao-Bench \cite{Zhang2023EvaluatingTP} collected questions from the 2010-2022 Chinese national college entrance examination papers, including 1,781 objective questions and 1,030 subjective questions, and constructed a framework for assessing the language comprehension and logical reasoning ability of LLMs. Microsoft has released a new benchmark test, AGIEval \cite{zhong2023agieval}, by selecting 20 official, public, high-standard exams, including general university entrance exams (Chinese national college entrance examination and the U.S. SAT), law school entrance exams, maths competitions, bar exams, national civil service exams, and more. Sun et. al. \cite{sun2023safety} proposed a benchmark on LLM security evaluation that encompasses a variety of typical security scenarios and command attack prompts for evaluating the security of LLMs. Sixteen different NLP tasks for medical scenarios are converted into prompt-based language generation tasks, constituting the first LLM evaluation benchmark for Chinese medical scenarios PromptCBLUE \cite{zhuwei2023promptcblue}.

Based on the existing work, it can be seen that the evaluation method for LLMs mainly involves constructing evaluation questions using existing topics or NLP-related tasks and evaluating LLMs based on their answers and responses. There are also approaches to dialogue with LLMs and manual evaluation through crowdsourcing. However, different LLMs learn different datasets during training, and for dataset $ \mathcal{D}$, LLM \textit{a} may have been learned while LLM \textit{b} may not have been learned. Therefore, there are risks of unfairness in evaluation tasks constructed based on pre-existing topics or datasets. Manual evaluation approaches can improve fairness to some extent, but the inability to automate evaluations leads to lower efficiency.

In order to evaluate the reasoning ability of LLMs more fairly, we would like to randomly automate the construction of evaluation tasks about LLMs. To achieve this goal, we chose to construct tasks to evaluate LLMs based on structurally flexible graph data. We first abstract the basic graph schema based on the existing social networks. After that, we randomly construct task graph templates based on the abstracted graph schema and construct LLM evaluation tasks based on the constructed graph templates.

our main contributions are as follows:

\begin{itemize}
    \item The first to consider a graph-based approach to evaluating LLMs.
    \item Proposing a fair method, named \textbf{\uline{Ro}cking \uline{Car}} (RoCar), for evaluating LLM's reasoning and memory skills.
\end{itemize}

In the next section, we will specifically describe the proposed LLM evaluation method.

\section{Methodology}
\label{methodology}
We think it is important to ensure that all tested LLMs have the same knowledge of the evaluation tasks during the training process for fairly evaluating the ability of LLMs. However, since the scale of the training datasets for LLM is often very large, it is difficult to determine whether a model has learned the evaluation tasks. In addition, evaluation tasks designed for a particular set of LLMs lack generalisability and cannot cope with LLM updates. Therefore, a naive idea to ensure that each LLM learns the same about the evaluation tasks is to ensure that all LLMs have not learned the evaluation tasks. Based on this idea, we use graph data with a flexible structure to construct the evaluation tasks, to improve the randomness and diversity of the tasks, and to avoid the evaluation task being learned over by LLMs. Therefore, we constructed the LLM evaluation tasks based on the social network graph and then proposed the RoCar method for evaluating the reasoning and memory abilities of LLMs. The RoCar consists of three main parts, i.e., basic graph schema extraction, generating the task graph, and constructing the evaluation tasks.
\subsection{Abstracting Basic Graph Schema}
We first extracted 1,144 relationship types from the social network graph collated by Liu \cite{liu2018chinesepersonrelationgraph}. We then removed second-order and higher relationships from these relationships (e.g., <maternal grandfather>, which can be represented as <mother, father>) and retained only first-order relationships (e.g., <son>, <older brother>, <father>, etc.). In addition, we removed hostile relationships (e.g. <love rival>, <enemy>, etc.). After that, we summarised some specific relationships and finally got 27 relationship types.

After obtaining these 27 relationship types, we further labeled the information corresponding to these relationship types to form the basic relationship graph schema. We labeled the gender, order, and direction of the relationships. The whole relationship types and their related information are shown in Table \ref{tab:relationship}.

\begin{table}
    \centering
    \caption{Basic Graph Schema of Relationships}
    \begin{tabular}{c|ccccc}
    \toprule
    No. &Head     &  Tail    &Type    &Order   &Direction  \\
    \midrule
    1    &2     & 2&  student &+  &1  \\
    2    &2   &   2   &teammate   &0  &2  \\
    3    &1 & 2 & son  & +   & 1 \\
    4    &0 & 2 & daughter  & +   & 1 \\
    5    &2 & 2 & friend  & 0   & 2 \\
    6    &0 & 2 & younger sister  & +   & 1 \\
    7    &2 & 2 & colleague  & 0   & 2 \\
    8    &1 & 2 & father  & 0   & 1 \\
    9    &0 & 1 & wife  & 1/- & 1 \\
    10    &2 & 2 & subordinate  & +/- & 1 \\
    11    &1 & 0 & boyfriend & 1/- & 1 \\
    12    &2 & 2 & leader  & +/- & 1 \\
    13    &1 & 2 & younger brother  & +   & 1 \\
    14    &2 & 2 & teacher  & +   & 1 \\
    15    &1 & 2 & older brother  & +   & 1 \\
    16    &1 & 2 & sworn younger brother & +/- & 1 \\
    17    &0 & 2 & sworn elder sister & +/- & 1 \\
    18    &0 & 1 & girlfriend & 1/- & 1 \\
    19    &0 & 2 & mother  & 0   & 1 \\
    20    &1 & 2 & sworn elder brother & +/- & 1 \\
    21    &0 & 2 & sworn younger sister & +/- & 1 \\
    22    &1 & 2 & godson & +/- & 1 \\
    23    &0 & 2 & goddaughter & +/- & 1 \\
    24    &1 & 2 & godfather  & +/- & 1 \\
    25    &0 & 2 & godmother  & +/- & 1 \\
    26    &0 & 2 & older sister  & +   & 1 \\
    27    &1 & 0 & husband  & 1/- & 1 \\
    \bottomrule
    \end{tabular}
    \label{tab:relationship}
\end{table}

In Table \ref{tab:relationship},  which contains 5 columns, $ head$ denotes the gender of the head node of the relationship, $ tail$ denotes the gender of the tail node of the relationship, $ type$ denotes the type of the relationship, $ order$ denotes the order of this relationship, and $ direction$ denotes the direction of the relationship. The meaning of the symbols in Table \ref{tab:relationship} is shown in Table \ref{tab:meanings}.

\begin{table}[]
    \centering
    \caption{Meanings of Symbols in Table \ref{tab:relationship}}
    \begin{tabular}{c|cl}
    \hline
    \multicolumn{1}{c|}{Attribute}                         & Symbol               & Meaning                                                                                                                                                                                                                                             \\ \hline
    \multicolumn{1}{c|}{\multirow{3}{*}{Head or Tail gender}} & 0                    & Female.                                                                                                                                                                                                                                             \\
    \multicolumn{1}{c|}{}                                  & 1                    & Male.                                                                                                                                                                                                                                               \\
    \multicolumn{1}{c|}{}                                  & 2                    & Both gender.                                                                                                                                                                                                                                           \\ \hline
    \multicolumn{1}{c|}{\multirow{4}{*}{Order}}            & 0                    & \begin{tabular}[c]{@{}l@{}}Concepts that do not require order, such as father, do not require \\ discussion of order under naive values.\end{tabular}                                                                                               \\
    \multicolumn{1}{c|}{}                                  & 1                    & \begin{tabular}[c]{@{}l@{}}Order is required but only one at present. For example, in the case \\ of the wife relationship, there can be multiple ex-wives, but only \\ one current one (considering the \textbf{rule} in China).\end{tabular} \\
    \multicolumn{1}{c|}{}                                  & +                    & \begin{tabular}[c]{@{}l@{}}The order of the current relationships, e.g. the relationship of brother \\ can have multiple brother relationships such as second brother, third \\ brother, etc.\end{tabular}                                          \\
    \multicolumn{1}{c|}{}                                  & -                    & A former relationship order, such as ex-girlfriends.                                                                                                                                                                                                \\ \hline
    \multicolumn{1}{c|}{\multirow{2}{*}{Direction}}        & 1                    & \begin{tabular}[c]{@{}l@{}}The direction of the relationship is from head to tail, e.g. \textit{h}(ead) is \\ the father of \textit{t}(ail).\end{tabular}                                                                                                             \\
    \multicolumn{1}{c|}{}                                  & 2                    & Relationships go both ways. For example, \textit{h}(ead) and \textit{t}(ail) are colleagues.                                                                                                                                                                          \\ \hline
    \multicolumn{1}{l}{}                                   & \multicolumn{1}{l}{} &                                                                                                                                                                                                                                                     \\
    \multicolumn{1}{l}{}                                   & \multicolumn{1}{l}{} &                                                                                                                                                                                                                                                     \\
    \multicolumn{1}{l}{}                                   & \multicolumn{1}{l}{} &                                                                                                                                                                                                                                                     \\
    \multicolumn{1}{l}{}                                   & \multicolumn{1}{l}{} &                                                                                                                                                                                                                                                     \\
    \multicolumn{1}{l}{}                                   & \multicolumn{1}{l}{} &     
    \end{tabular}
    \label{tab:meanings}
\end{table}

Each relationship in Table \ref{tab:relationship} represents a basic schema of the task graph. It contains the gender of the head and tail nodes corresponding to a first-order relation, the type of relation, the order of the relation, and the direction of the relation.

\subsection{Template Definition and Randomised Social Network Graph Generation}

With the abstracted basic schema, we can construct a random task graph. First, we randomly select a set of 27 basic schemas for constructing the task graph, denoted by $ \mathcal{B}$, which contains \textit{n} repeatable basic schemas. 
For example, suppose there is a need to construct a task graph using \textbf{three} basic schemas to evaluate LLMs and we randomly selected three basic schemas as shown in Table \ref{tab:example}.

\begin{table}
    \centering
    \caption{An Example of A Set of Basic Schemas}
    \begin{tabular}{c|ccccc}
    \toprule
    No. & Head & Tail & Type          & Order & Direction \\
    \midrule
    1   & 2    & 2    & student  & +     & 1         \\
    2   & 1    & 2    & son      & +     & 1         \\
    3   & 0    & 2    & daughter & +     & 1        \\
    \bottomrule
    \end{tabular}
    \label{tab:example}
\end{table}

\begin{figure}
    \centering
    \includegraphics[width=1\textwidth]{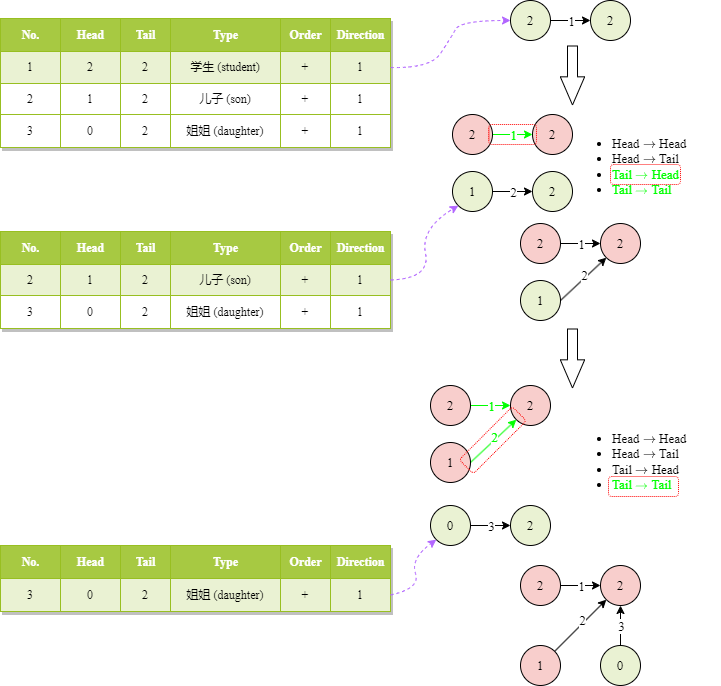}
    \caption{The process of constructing a task graph.The figure contains two columns; the first column shows the process of decreasing the basic schema as it is inserted into the figure, while the second column shows the dynamic process of constructing an evaluation task graph from the basic schema. The \textcolor[HTML]{EAF2D3}{green} nodes in the figure indicate newly inserted basic schemas and the \textcolor[HTML]{F8CFCD}{red} nodes indicate constructed task graphs. The \textcolor[HTML]{00FF00}{green} relationships and splicing methods indicate optional relationships or splicing methods, and the \textcolor[HTML]{FF0000}{red} boxes indicate randomly selected relationships or splicing methods.}
    \label{fig:task-graph}
\end{figure}

For this set of basic schemas, we first perform a random ordering, and the sorted order is shown in Table \ref{tab:example}. After that, we build the task graph in this order. For each selected basic schema, we randomly chose a relationship in the constructed task graph for splicing. There are a total of four possibilities for each splice, and all splice forms are shown in Table \ref{tab:splice}, where we randomly select one form of splice from the feasible ones. Based on the example in Table \ref{tab:example}, the process of constructing the task graph is shown in Figure \ref{fig:task-graph}.

\begin{table}
    \centering
    \caption{Splicing Methods}
    \begin{tabular}{c||l|l}
    \hline
    No. & Linking Method & \multicolumn{1}{c}{Meaning}                                                                                                                                                             \\ \hline
    1   & Head $ \to$ Head   & \begin{tabular}[c]{@{}l@{}}The head node of the current basic schema is set to be the same as \\ the head node of a randomly selected relation in the existing task graph.\end{tabular} \\
    2   & Head $ \to$ Tail   & \begin{tabular}[c]{@{}l@{}}The head node of the current basic schema is set to be the same as \\ the tail node of a randomly selected relation in the existing task graph.\end{tabular} \\
    3   & Tail $ \to$ Head   & \begin{tabular}[c]{@{}l@{}}The tail node of the current basic schema is set to be the same as \\ the head node of a randomly selected relation in the existing task graph.\end{tabular} \\
    4   & Tail $ \to$ Tail   & \begin{tabular}[c]{@{}l@{}}The tail node of the current basic schema is set to be the same as \\ the tail node of a randomly selected relation in the existing task graph.\end{tabular} \\ \hline
    \end{tabular}
    \label{tab:splice}
\end{table}

Based on the process shown in Figure \ref{fig:task-graph}, a task graph for evaluation can be constructed based on the basic model exemplified in Table \ref{tab:example}.

\subsection{Evaluation Tasks Construction}
To facilitate the evaluation of LLMs, we transformed the task graph constructed above into natural language prompts and questions. In addition to randomly constructing the task graphs, we constructed surrogate libraries containing names and genders to ensure fairness, e.g., ``{name: Xiaohong), gender: female}''. 

We populate each node with a name surrogate by gender in order of relationship and randomize the ``+'' and ``-'' in it to the corresponding values. Afterward, we converted each relation in the task graph into a natural language prompt according to a predefined template. The natural language prompts corresponding to each type of relationship transformation are shown in Table \ref{tab:prompts}.

\begin{table}
    \centering
    \caption{Prompts of Basic Schemas}
    \begin{tabular}{c|l}
    \hline
    No. & Prompt                                                        \\ \hline
    1   & Xiaohong is Xiaoming's third student.             \\
    2   & Xiaohong and Xiaoming are teammates.                 \\
    3   & Xiaogang is Xiaoming's third son.                   \\
    4   & Xiaohong is Xiaoming's third daughter.              \\
    5   & Xiaohong and Xiaoming are friends.                   \\
    6   & Xiaohong is Xiaoming's third younger sister.         \\
    7   & Xiaohong and Xiaoming are colleagues.                \\
    8   & Xiaoming is Xiaohong's father.                       \\
    9   & Xiaohong is Xiaoming's wife.                         \\
    10  & Xiaohong is Xiaoming's subordinate.                  \\
    11  & Xiaoming is Xiaohong's boyfriend.                   \\
    12  & Xiaohong is Xiaoming's leader.                       \\
    13  & Xiaoming is Xiaohong's third younger brother.        \\
    14  & Xiaohong is Xiaoming's third teacher.             \\
    15  & Xiaoming is Xiaohong's third older brother.          \\
    16  & Xiaoming is Xiaohong's third sworn younger brother. \\
    17  & Xiaohong is Xiaoming's third sworn elder sister.    \\
    18  & Xiaohong is Xiaoming's girlfriend.                  \\
    19  & Xiaohong is Xiaoming's mother.                       \\
    20  & Xiaoming is Xiaohong's third sworn elder brother.   \\
    21  & Xiaohong is Xiaoming's third sworn younger sister.  \\
    22  & Xiaoming is Xiaohong's third surrogate son.            \\
    23  & Xiaohong is Xiaoming's third surrogate daughter.        \\
    24  & Xiaoming is Xiaohong's third informal godfather.             \\
    25  & Xiaohong is Xiaoming's third informal godmother.             \\
    26  & Xiaohong is Xiaoming's older sister.                 \\
    27  & Xiaoming is Xiaohong's husband.                      \\ \hline
    \end{tabular}
    \label{tab:prompts}
\end{table}

To further ensure the fairness of the evaluation method, we also organized the designations corresponding to the basic and second-order relations into natural language prompts. These prompts are fed into the LLMs prior to testing to ensure that the LLMs being tested are aware of the basic designations in the social network. In addition, we have organized the rules for appellative derivation into natural language prompts as well.

After LLMs had learned the basic prompts, we asked LLMs questions using questions constructed based on the task graph and evaluated LLMs based on the correctness of the answers they responded to. The question was constructed by randomly selecting two nodes from the task graph and forming a natural language question based on the names corresponding to the nodes, asking LLMs about the relationship or designation between the two people. There are four main forms of problems:

\begin{itemize}
    \item What's the relationship between Xiaoming and Xiaohong?
    \item What should Xiaoming call Xiaohong?
    \item Is there a mother-son relationship between Xiaoming and Xiaohong?
    \item Should Xiaoming call Grandma Xiaohong?
\end{itemize}

We next present the results of the experiments in Section \ref{experiments}.

\section{Experiments}
\label{experiments}
Our work focuses on evaluating LLMs in terms of both reasoning and memory capabilities. The files can be found at \url{https://github.com/NEU-DataMining/RoCar}
\subsection{Design of Experiments}
\label{design}
Based on the introduction in Section \ref{methodology}, we randomly constructed a task graph and a series of evaluation tasks.

For the evaluation of reasoning ability, we divided the evaluation tasks into groups according to the distance between the two people in the evaluation tasks on the task graph. Afterward, we randomly screened one of the evaluation tasks at each distance. Finally, we get one evaluation task each with distances ranging from \textbf{2} to \textbf{5}. We asked each LLM separately with evaluation tasks for each distance from 2 to 5 and judged the correctness of the answers given by the LLMs. Finally, we scored the reasoning ability of LLMs based on all the results, and the longer the distance the greater the weight of the evaluation task. The scoring criteria are shown in (\ref{eq:reson_score}):

\begin{equation}
    score_r = \frac{\sum^5_{i=2} p_i \times i}{\sum^5_{i=2}} \ \ \ \ \ \ , p_i = \begin{cases}
        1, & \text{the\ result\ is\ correct,} \\
        0.5,    & \text{the result is correct but the logic is wrong,} \\
        0, & \text{the\ result\ is\ wrong.}
    \end{cases}
    \label{eq:reson_score}
\end{equation}

For the memory capability evaluation, we informed the LLMs about the task graph in multiple steps, from 1 to 5 steps for the experiment. Then, we selected evaluation tasks with distances of 1 and 2 to test LLMs. The higher the number of steps, the higher the weight of the corresponding test result. In addition, the weight of the test results for tasks with a distance of \textbf{2} is higher than the test results for tasks with a distance of \textbf{1}. The scoring criteria are shown in (\ref{eq:memory_score}):

\begin{equation}
    score_m = 0.25 \times \frac{\sum^5_{i=1} p^{(1)}_i \times i}{\sum^5_{i=1}}\ + \ 0.75 \times \frac{\sum^5_{i=1} p^{(2)}_i \times i}{\sum^5_{i=1}} \ \ \ \ \ \ , p_i = \begin{cases}
        1, & \text{the\ result\ is\ correct,} \\
        0.5,    & \text{relationship is correct but result is wrong,} \\
        0, &\text{the\ result\ is\ wrong.}
    \end{cases}
    \label{eq:memory_score}
\end{equation}

\subsection{Reasoning Capability}
\label{reasoning}
Based on the introduction in Section \ref{design}, we constructed a task graph shown in Figure \ref{fig:task_graph} and tested the reasoning capability of some open-use LLMs. The results are shown in Table \ref{tab:score_r}:

\begin{figure}
    \centering
    \includegraphics[width=1\textwidth]{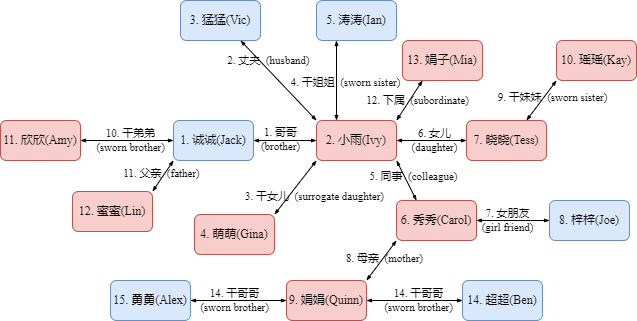}
    \caption{Task graph. \textcolor[HTML]{F8CFCD}{Red} nodes indicate females, \textcolor[HTML]{DBE9FC}{blue} nodes indicate males, arrows indicate the relationship between the two, and ordinal numbers indicate the order in which the task graph was constructed. }
    \label{fig:task_graph}
\end{figure}

\begin{table}
    \centering
    \caption{Reasoning Scores}
    \begin{tabular}{c|cccccc}
    \hline
    LLM      & ChatGPT & ChatGLM & Ernie-Bot & Spark-Desk & Claude \\
    \hline
    score\_r & 7.14     & \uline{78.57}     & 42.86       & \textbf{100.00}        & \uline{78.57}   \\
    \hline
    \end{tabular}
    \label{tab:score_r}
\end{table}

\subsection{Memory Capability}
Based on the introduction of Section \ref{design} and the task graph in Section \ref{reasoning}, we tested the memory capability of LLMs. The results are shown in Table \ref{tab:score_m}.

\begin{table}
    \centering
    \caption{Memory Capability Scores}
    \begin{tabular}{c|cccccc}
    \hline
    LLM      & ChatGPT & ChatGLM & Ernie-Bot & Spark-Desk & Claude \\
    \hline
    score\_m & \uline{68.33}     & 8.33     & 15.00      & 29.17        & \textbf{88.33}   \\
    \hline
    \end{tabular}
    \label{tab:score_m}
\end{table}

\subsection{Analysis of Results}
We tested a number of open-use LLMs using a randomly constructed social relation graph. Note that during the experimental process of the reasoning ability test, due to ChatGLM's poor ability to memorize the task graph in multiple rounds, we re-informed it about the task graph before asking questions about each evaluation task in order to ensure that the experiments proceeded smoothly.

In the experiment, Claude had the best overall performance, which is in line with our general perception. However, ChatGPT did not perform particularly well in the reasoning ability test experiment. We hypothesize that this may be due to its lower Chinese comprehension and understanding of social networks in Chinese culture. In addition, we conducted only one set of randomized experiments during the testing process, which may have led to results that were subject to serendipity.

\section{Conclusion}
In this paper, we propose an LLM evaluation method based on graph-structured data, RoCar. We constructed the basic schema, the library of surrogates, and the derivation rules for social network graphs. Task graphs for evaluation tasks can be constructed randomly based on basic schemas and pronominal libraries. Evaluation tasks can be constructed based on the task graph after organizing it into natural language prompts. Afterward, we used these evaluation tasks to test LLMs' reasoning and memory capabilities, respectively. There is a high degree of randomization in our proposed evaluation method, which can greatly improve the fairness of the evaluation.

\section{Future Work}
Although the method we proposed does a good job of improving the fairness of the evaluation, there is still a lot of room for improvement in this method. In the future, we can improve our work in the following areas:

\begin{itemize}
    \item Expand the number of relationship types. Combine social networks with other types of graphs to construct more realistic and complex task graphs.
    \item Adding relationships that exist in reality but are not in line with naive values and expanding the related evaluation tasks to evaluate LLMs in terms of values alignment, bias, harmfulness, etc.
    \item Evaluate more LLMs.
    \item Conduct multi-group randomized experiments to further improve the fairness of the evaluation.
    \item Enrich the types and formats of prompts and validate the sensitivity of different LLMs to the prompts.
    \item Constructing multilingual task graphs and evaluation tasks.
\end{itemize}


\bibliographystyle{unsrt}  
\bibliography{references}

\end{document}